\documentclass[acmtog]{acmart} 
\acmSubmissionID{1719}

\usepackage{multirow}
\usepackage{bbding}
\usepackage{soul}
\usepackage{colortbl}

\usepackage{pifont}
\usepackage{enumitem}
\usepackage[ruled]{algorithm2e} 
\usepackage{algorithmic} 

\newcommand{\best}[1]{\textbf{#1}}
\newcommand{\second}[1]{\underline{#1}}

\SetAlFnt{\small}
\SetAlCapFnt{\small}
\SetAlCapNameFnt{\small}
\SetAlCapHSkip{0pt}

\acmJournal{TOG} 

\begin{document}
\title{
DvD: Unleashing a Generative Paradigm for Document Dewarping via 
Coordinates-based Diffusion Model}

\author{Weiguang Zhang}
\orcid{0009-0005-8783-0326}
\affiliation{
\institution{Xi'an Jiaotong-Liverpool University} 
\city{Suzhou}
\postcode{215123}
\country{China}
}
\affiliation{
\institution{University of Liverpool}
\city{Liverpool}
\country{United Kingdom}
}
\email{Weiguang.Zhang21@student.xjtlu.edu.cn}

\author{Huangcheng Lu} 
\orcid{0009-0002-9808-8823}
\affiliation{
\institution{Xi'an Jiaotong-Liverpool University} 
\city{Suzhou}
\postcode{215123}
\country{China}
}
\affiliation{
\institution{University of Liverpool}
\city{Liverpool}
\country{United Kingdom}
}
\email{huangcheng.lu22@student.xjtlu.edu.cn}

\author{Maizhen Ning} 
\orcid{0000-0002-8842-4187}
\affiliation{
\institution{Xi'an Jiaotong-Liverpool University} 
\city{Suzhou}
\postcode{215123}
\country{China}
}
\affiliation{
\institution{University of Liverpool}
\city{Liverpool}
\country{United Kingdom}
}
\email{maizhen.ning16@student.xjtlu.edu.cn}
\author{Xiaowei Huang} 
\orcid{0000-0001-6267-0366}
\affiliation{
 \institution{University of Liverpool}
 \city{Liverpool}
 \country{United Kingdom}
}

\author{Wei Wang} 
\orcid{0000-0002-0707-8076}
\affiliation{
 \institution{Xi'an Jiaotong-Liverpool University}
 \city{Suzhou}
 \postcode{215123}
 \country{China}}

\author{Kaizhu Huang}
\orcid{0000-0002-3034-9639}
\affiliation{
 \institution{Duke Kunshan University}
 \city{Suzhou}
 \postcode{215316}
 \country{China}}

\author{Qiufeng Wang\textsuperscript{†}}
\authorsaddresses{† Corresponding author: Qiufeng Wang, \href{mailto:Qiufeng Wang}{Qiufeng.Wang@xjtlu.edu.cn}}
\orcid{0000-0002-0918-4606}
\affiliation{
 \institution{Xi'an Jiaotong-Liverpool University}
 \city{Suzhou}
 \postcode{215123}
 \country{China}}
\email{Qiufeng.Wang@xjtlu.edu.cn}

\renewcommand\shortauthors{Zhang, W. et al}

\begin{abstract}
Document dewarping aims to rectify deformations in photographic document images, thus improving text readability, which has attracted much attention and made great progress, but it is still challenging to preserve document structures. Given recent advances in diffusion models, it is natural for us to consider their potential applicability to document dewarping. However, it is far from straightforward to adopt diffusion models in document dewarping due to their unfaithful control on highly complex document images (e.g., 2000$\times$3000 resolution). 
In this paper, we propose DvD, the first generative model to tackle document \textbf{D}ewarping \textbf{v}ia a \textbf{D}iffusion framework. To be specific, DvD introduces a coordinate-level denoising instead of typical pixel-level denoising, generating a mapping for deformation rectification. In addition, we further propose a time-variant condition refinement mechanism to enhance the preservation of document structures. In experiments, we find that current document dewarping benchmarks can not evaluate dewarping models comprehensively. To this end, we present AnyPhotoDoc6300, a rigorously designed large-scale document dewarping benchmark comprising 6,300 real image pairs across three distinct domains, enabling fine-grained evaluation of dewarping models. Comprehensive experiments demonstrate that our proposed DvD can achieve state-of-the-art performance with acceptable computational efficiency on multiple metrics across various benchmarks, including DocUNet, DIR300, and AnyPhotoDoc6300. The new benchmark and code will be publicly available at https://github.com/hanquansanren/DvD. 

\end{abstract}




\maketitle

\section{Introduction}
\label{sec:intro}
With the ubiquity of smartphones, taking photos to digitize documents has become an increasingly convenient and common practice. However, compared with the scanned documents, document images captured in daily scenes often suffer from poor readability caused by severe geometric deformations (e.g., folds, curves, crumples, etc.). These deformations hinder both human readability and specialist optical character recognition (OCR) engines, even multimodal large language models (MLLMs)~\cite{Scius_ocr_2024}. 
To improve readability, document dewarping task emerges to restore flat documents before downstream document digitization pipelines (e.g., Layout analysis, Text spotting)~\cite{li2025docsam, wan_2024_multidetect}, achieving comparable OCR performance to flat counterparts.
\begin{figure}[t]
	\centering
	\includegraphics[width=\linewidth]{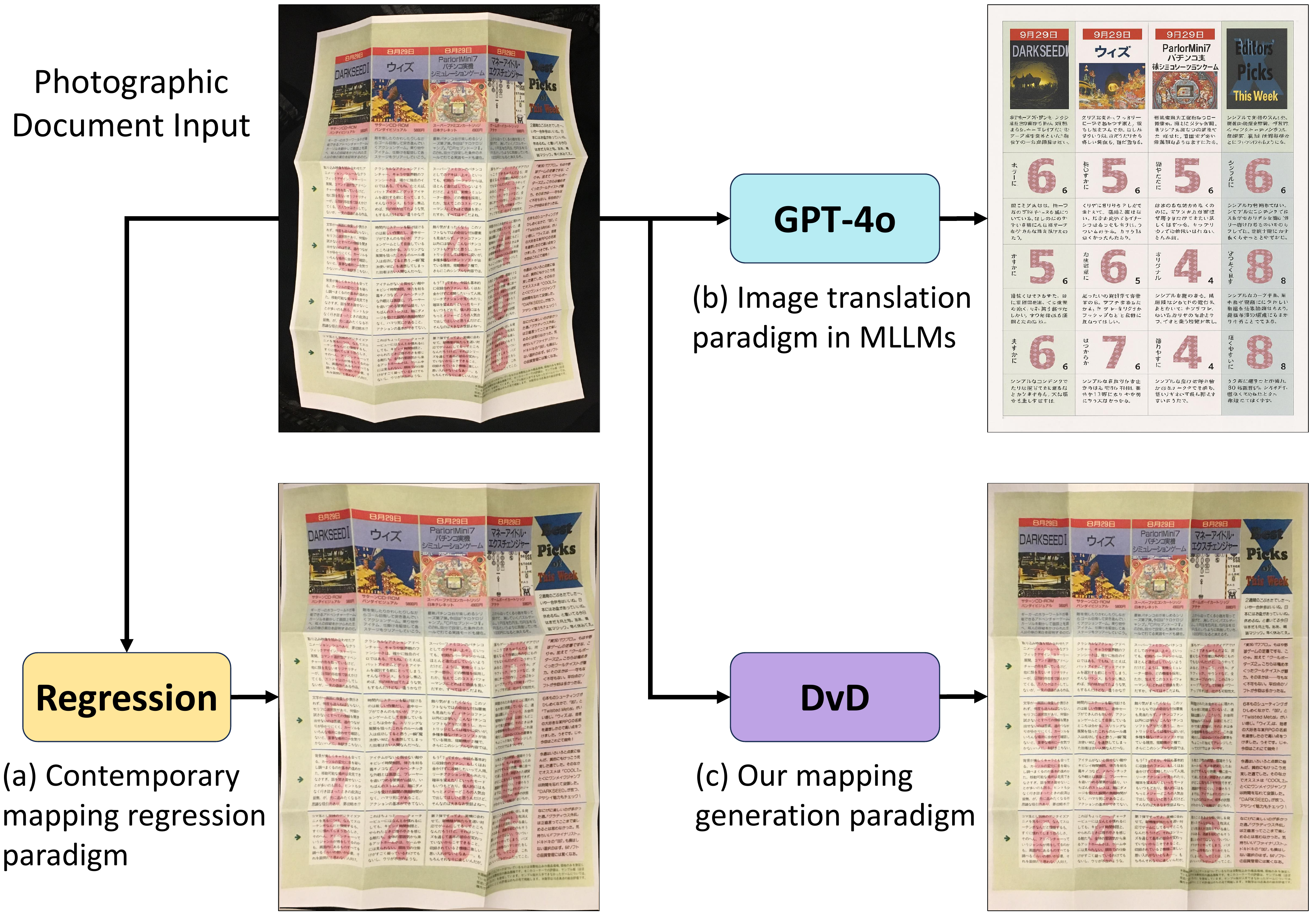}
	\caption{Comparisons to the existing paradigm for document
dewarping. Unlike the existing paradigms (a) and (b) struggle with either precise structural preservation or faithful content generation, our DVD (c) can yield flat document images with precise yet faithful content through a mapping generation paradigm.
}
\label{fig:head_img}
\end{figure} 



Document dewarping has evolved through two distinct periods. The early model-based methods~\cite{Kanungo_1993, cao2003cylindrical, RN19} basically follow a "reconstruct-then-dewarp" paradigm, which is typically limited by hardware dependence or surface assumptions.
Subsequent data-driven methods~\cite{ma2018docunet, li2019document, das2019dewarpnet} have shifted toward a mapping regression paradigm to model the condition probability. Specifically, given a large-scale photographic document dataset, these methods predominantly train neural networks to directly regress mapping (e.g., backward mapping, forward mapping) for dewarping. Accordingly, they also transfer the research attention to high-fidelity training data~\cite{zhang_2024_docregistration, verhoeven2023neural, zhang2024reg} and better network architecture for feature extraction~\cite{li2023foreground, Feng2022geodocnet}. 

Albeit accomplishing notable advancements, the contemporary mapping regression paradigm is burdened by its intrinsic discriminative nature, which lacks the capacity to explicitly capture the underlying data distribution, resulting in imprecise structural preservation (See Fig.~\ref{fig:head_img} (a)).
Recent diffusion models have demonstrated the viability of employing a generative paradigm to solve discriminative tasks~\cite{ravishankar_2024_scaling, nam_2024_diffmatch, luo2024flowdiffuser}. By learning a progressive denoising process to generate samples conforming to the training data distribution, these models introduce a generative task modeling to learn more comprehensive distributions.
Most recently, MLLMs (e.g., Gemini 2.5, GPT-4o) have exhibited remarkable natural image generation capabilities~\cite{yan2025gpt-imgeval}, but failed to preserve document structures in document dewarping (e.g., Fig.~\ref{fig:head_img} (b)). We attribute this to the fact that it’s very difficult to directly employ an image translation
paradigm to obtain faithful control in highly complex document images. In light of these insights, we propose one intriguing question: \textit{Can generative models be effective for document dewarping?}



To answer this question, we present DvD, the first effort to unleash a generative dewarping model built on a coordinates-based denoising diffusion framework. Instead of typical pixel-level denoising, we introduce a coordinate-level denoising process, where DvD learns how to progressively generate a series of latent variables to characterize the mapping for deformation rectification. We argue that such a mapping generation paradigm not only explicitly fosters deformation-aware modeling but also avoids the difficulty of high-resolution image generation. 
To further enhance the structural preservation, DvD also incorporates a time-variant condition refinement mechanism to leverage intermediate dewarping results. Diverging from typical diffusion models guided by a time-fixed condition, our proposed time-variant mechanism enables intermediate-aware dynamic guidance in the denoising generation process.

To make the evaluation of dewarping models fairly, we collect most of publicly available document dewarping benchmarks as shown in  Table~\ref{tab:unwarping_datasets}. We find that these benchmarks typically suffer from three critical limitations that impede a comprehensive evaluation of dewarping models, including restricted coverage of real-world scenarios, a small dataset scale, and deficient annotation of domains. 
These limitations might have led to evaluation ambiguity, restricting the model's application in real-world scenarios. To this end, we construct a fine-grained and large-scale benchmark AnyPhotoDoc6300, which contains 6,300 real-world photographic image document pairs, rigorously organized across three distinct domains, enabling fine-grained evaluation of dewarping models. In addition to the benchmark, we extend the evaluation metrics. We find that existing methods commonly utilize off-the-shelf OCR engines to measure the Edit Distance (ED) and Character Error Rate (CER) as OCR metrics~\cite {das2019dewarpnet}. 
We identify that there is still no exploration about whether the dewarped document can attain equivalent text readability to its flat counterpart for MLLMs.
To fill the blank, we propose two MLLM-based OCR metrics (MMCER and MMED) in this work.



In summary, our main contributions are four-fold:
\begin{itemize}[topsep=1pt,leftmargin=*] 
\item Pioneering a paradigm shift, we present DvD, the first generative model to tackle document dewarping via a coordinates-based diffusion framework. Unlike typical pixel-level denoising to generate dewarped images directly, we operate coordinate-level denoising to generate coordinate mappings for dewarping.
\item We introduce a time-variant condition refinement mechanism that dynamically leverages intermediate dewarping results as guidance to enhance the preservation of document structures. 
\item To offer a comprehensive evaluation of document dewarping models, we construct a fine-grained benchmark dataset AnyPhotoDoc 6300, which contains 6,300 real-world photographic image document pairs, rigorously organized across three distinct domains.
\item Extensive experiments demonstrate state-of-the-art dewarping performance with acceptable computational efficiency. 
\end{itemize}

\begin{table}[t]
\centering
\caption{Comparison of document dewarping benchmarks. \ding{55} symbolizes that this benchmark doesn't explicitly distinguish this domain. \textbf{Scenes:} Multiple scenarios (Mul), Real (R), Synthetic (S). \textbf{Domains:} Layouts Category (L), Environment Lighting (E), Capture Angles (A). }
\label{tab:unwarping_datasets}
\resizebox{\linewidth}{!}{
\begin{tabular}{lccccc}
\toprule
\multirow{2.5}{*}{\textbf{Dataset}} &\multirow{2.5}{*}{\textbf{Scenes}} &\multirow{2.5}{*}{\textbf{Images}} &\multicolumn{3}{c}{\textbf{Domains}} \\
\cmidrule(lr){4-6}
 & & &\textbf{L} &\textbf{E} &\textbf{A}\\
\midrule
DocUNet~\cite{ma2018docunet} &Mul-R &$130\times2$  &\ding{55}  &\ding{55} &\ding{55}\\             
DIR300~\cite{Feng2022geodocnet}  &Mul-R &$300\times2$    &\ding{55}  &\ding{55} &\ding{55}\\
Inv3DReal~\cite{hertlein2023inv3d} &Invoice-R  &$360\times2$    &1  &3 &1\\
UVDoc~\cite{verhoeven2023neural} &Mul-S &$50\times2$ &\ding{55} &\ding{55} &\ding{55}\\
DocReal~\cite{yu_docreal_2024} &Chinese-R  &$200\times2$    &\ding{55}  &\ding{55} &\ding{55}\\
\textbf{Our AnyPhotoDoc6300} &\textbf{Mul-R} &$\textbf{6300}\times2$  &\textbf{8}  &\textbf{3} &\textbf{2}\\
\bottomrule
\multicolumn{6}{l}{\small $\bullet$ Noted that we don't list WarpDoc~\cite{xue2022fourier} and SP~\cite{li2023ladoc} }\\ 
\multicolumn{6}{l}{ due to open-source incompleteness.}
\end{tabular}}
\end{table}

\begin{figure*}[ht]
	\centering
	\includegraphics[width=\linewidth]{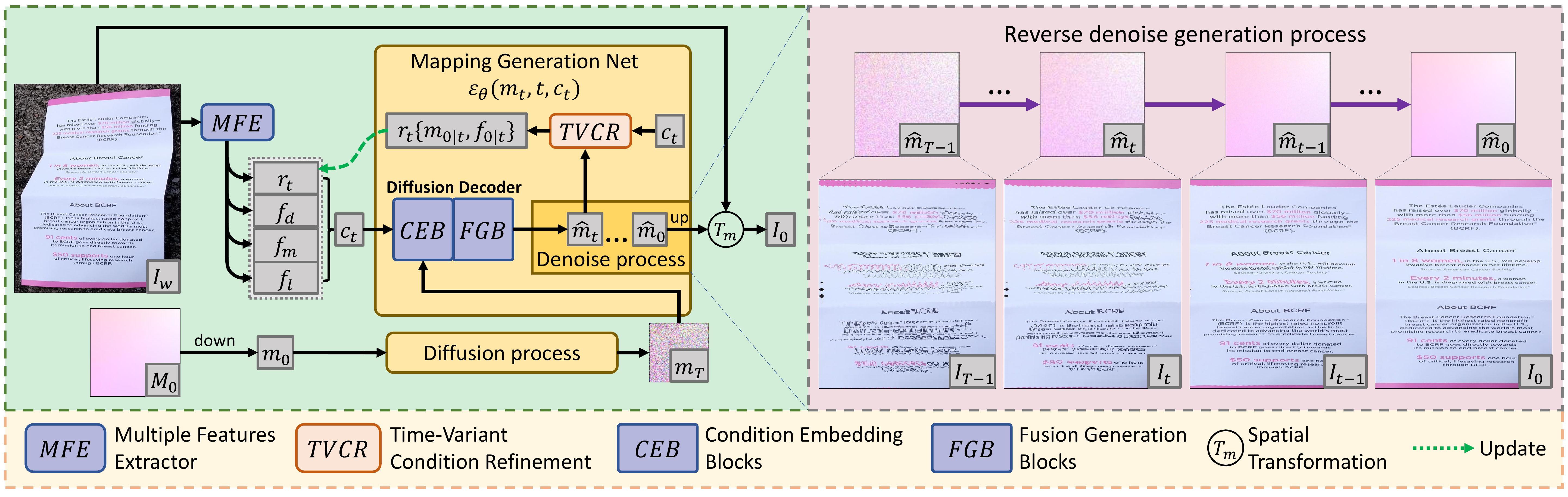}
	\caption{\textbf{Framework of the proposed DvD.} Given a warped document as input, DvD generates latent variables $m$ to characterize the mapping for deformation rectification. We leverage a compound condition $c_t$: the features for raw document images $f_d$, document foreground mask $f_m$, text-lines $f_l$, and the time-variant condition of the intermediate result $r_t$, respectively. We visualize the reverse denoise generation process in the right pink region.}
	\label{fig:img2}
\end{figure*}

\section{Related work}
\label{sec:related}
\subsection{Early Model-based Methods}
Early model-based methods basically adhere to a two-step "reconstruct-then-dewarp" paradigm. At the first step, for surface reconstruction, some methods utilize specialized hardware such as structured-light devices~\cite{RN26, RN21}, laser scanners~\cite{RN29}, multi-view camera systems~\cite{Ulges_2004, koo2009composition, RN11}, and proximal light sources~\cite{RN23}. Another methods bypass the hardware dependence by leveraging 2D visual cues (e.g., text lines~\cite{tian2011rectification}, local text orientation~\cite{RN18}, text blocks~\cite{RN16}, etc.) to estimate 3D geometry with parametric assumptions. 
In the second step, hand-crafted transformations for dewarping the surface to the plane are undertaken according to the reconstructed surface. For parameterized surfaces, typical transformations involve Generalized Cylinders Surface (GCS)~\cite{RN13, koo2009composition}, developable surfaces~\cite{RN19}, generalized ruled surfaces~\cite{Tsoi_2007}, and Non-Uniform Rational B-Splines (NURBS)~\cite{tan2005restoring}. For non-parameterized surfaces, techniques such as planar strips~\cite{RN20}, stiff mass-spring systems~\cite{RN29}, conformal mapping~\cite{RN11}, and sparse correspondence~\cite{RN18} are harnessed to tackle the mesh representation.
While effective in limited scenarios, these methods exhibited poor generalization to real-world warping patterns due to confined assumptions. Meanwhile, demanding hardware-dependent solutions also deviates from the prevalent habit of using mobile phones to digitize documents.

\subsection{Data-driven Methods}
Data-driven methods bring a shift toward a regression-based paradigm, significantly lessening reliance on both assumptions and hardware. Ma et al.~\citeyearpar{ma2018docunet} pioneer the application of deep neural networks to solve document dewarping by framing it as a mapping regression task. Subsequent DewarpNet~\cite{das2019dewarpnet} reformulates the traditional "reconstruct-then-dewarp" paradigm via two regression networks while collecting a high-quality synthetic dataset Doc3D using rendering software. CREASE~\cite{markovitz2020can} explores multi-modal warped document representations to strengthen the dewarping performance.
DocProj~\cite{li2019document} and PW-DewarpNet~\cite{das2021end} opt to dewarp documents in a patch-wise manner, benefiting the results at local details. DispFlow~\cite{xie2020dewarping} and DDCP~\cite{xie2021document} investigate the effectiveness of mapping and sparse control points as deformation representations, respectively. DocTr~\cite{feng2021doctr} replaces the convolutional network with a vision transformer, achieving significant performance boosts. 
RDGR~\cite{jiang2022RDGR}, DocGeoNet~\cite{Feng2022geodocnet}, and FTA~\cite{li2023foreground} leverage text-line features to keep textual content alignment. PaperEdge~\cite{RN58}, Marior~\cite{jiaxin2022marior}, and DocReal~\cite{yu_docreal_2024} devise two-stage networks, which first coarsely dewarp the mask of document, then refine details. 
FDRNet~\cite{xue2022fourier} introduces frequency-domain insight, designing an image-level loss based on high-frequency textures extracted from the Fourier transformation. UVDoc~\cite{verhoeven2023neural} exploits a novel data annotation pipeline through optical invisibility of ultraviolet ink to acquire pseudo-real training data. LA-Doc~\cite{li2023ladoc} constructs a synthetic dataset with large-scale deformation patterns via physical mass-spring systems and proposes a layout-aware dewarping network. Inv3D and DocMatcher~\cite{hertlein2025docmatcher,hertlein2023inv3d} present a template-based document dewarping using auxiliary template information. DocHFormer~\cite{zhou_dochformer_2025} introduces a novel shuffle transformer block to harmonize feature representation. DocRes~\cite{zhang_2024_docres} develops a unified model that integrates multiple low-quality document enhancement tasks. These methods concentrate on curating training datasets and devising regression networks, whereas their intrinsic discriminative nature is unexplored, which constrains the dewarping performance.

\section{Methodology}
\label{sec:method}


\begin{figure}[tp]
	\centering
	\includegraphics[width=\linewidth]{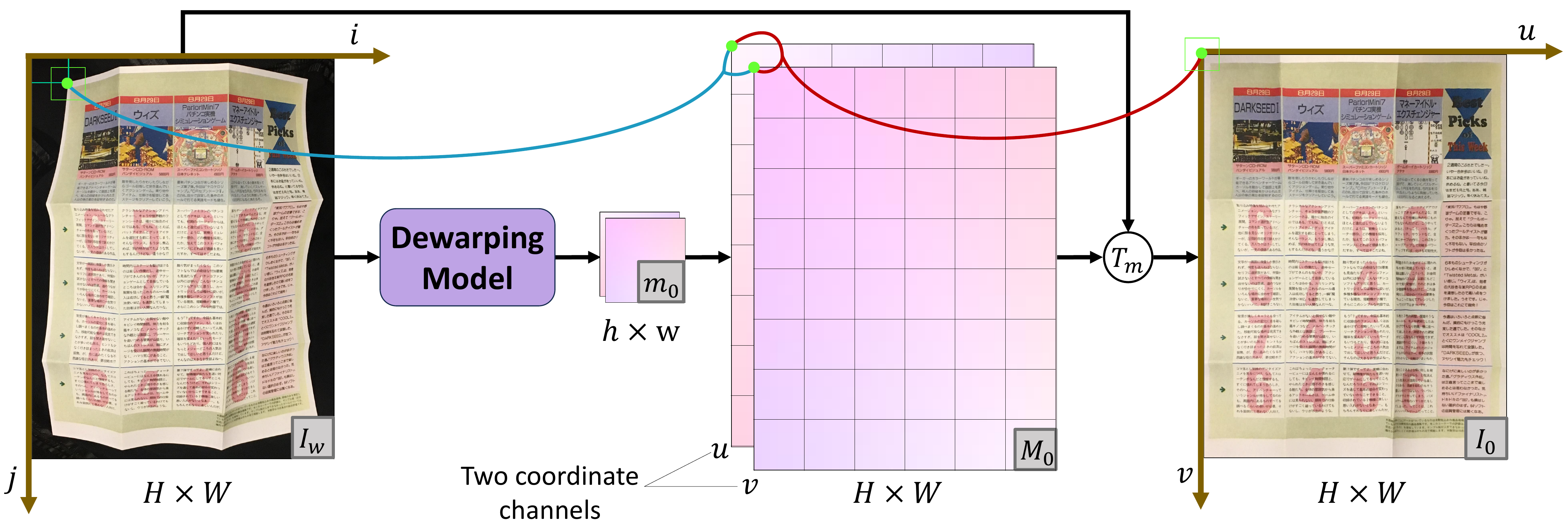}
	\caption{General document dewarping pipeline. We propose a dewarping model to predict a backward mapping for deformation rectification, consisting of two coordinates channels.}
\label{fig:head_dewarp}
\vspace{-10pt}
\end{figure}

\subsection{Document Dewarping Framework}
\label{sec:Framework}
As illustrated in Fig.~\ref{fig:head_dewarp}, given a warped photographic document image $I_w\in \mathbb{R}^{H\times W\times 3}$ as input, document dewarping aims to restore the flat document texture $I_0\in \mathbb{R}^{H\times W\times 3}$. In this work, we firstly propose a dewarping model to predict a small-scale coordinates mapping $m_0$, saving computational cost, which is then up-sampled to a normal backward mapping $M_0\in \mathbb{R}^{H\times W\times 2}$ (each value in $M_0$ represents the corresponding 2D coordinates in the input warped image $I_w$ as shown in Equ.~\ref{eq:def}), Next, we employ a dense spatial transformation $T_{m}$ to calculate the pixel values in $I_w$ according to the coordinates from $M_0$, ultimately obtaining $I_0$ as shown in Equ.~\ref{eq:def}. 
We formalize the backward mapping procedure as
\begin{equation}
\label{eq:def}
\begin{split}
(i,j)&=M_0(u,v),\\
I_0 (u,v)&= T_{m}(I_w(i,j)),
\end{split}
\end{equation}
where $(i,j)$ and $(u,v)$ represent the 2D spatial coordinates of $I_w$ and $I_0$, respectively. In the following, we will describe our dewarping model in details.


\subsection{Coordinates-Based Diffusion Model}
\label{sec:method31}
\subsubsection{Latent Diffusion Model in Coordinate Space.}
Directly generating $M_0$ explicit foster deformation-aware modeling to achieve better structural preservation, however both $M_0$ and $I_0$ actually contain equally high-resolution (e.g., 2000×3000), leading to high computational complexity. Inspired by the latent diffusion model~\cite{rombach2021highresolution}, we implement the diffusion and denoising processes only in a smaller coordinate space (i.e., $64\times64$), significantly reducing the training and inference costs. Figure~\ref{fig:img2} shows the overall framework of our proposed DvD model. Concretely, guided by a compound condition $c_t$, our DvD progressively generates a series of latent variables $m$ from a random Gaussian distribution. This paradigm relies on a tailored conditional denoising diffusion probabilistic framework, which typically contains both forward and reverse processes~\cite{ho_2020_ddpm, Song_2019_Generative}. We define the forward diffusion process for mapping as the Gaussian transition, s.t. $q(m_{t} | m_{t-1}) :=\mathcal{N}(\sqrt{1-\beta_t}m_{t-1}, \beta_t \mathbf{I})$, where ${\beta_t}$ is a predefined variance schedule. The resulting latent variable $m_t$ can be formulated as:
\begin{equation}
\label{eq:forward}
m_t = \sqrt{\bar{\alpha}_t}m_0 + \sqrt{1 - \bar{\alpha}_t}z, \quad z \sim \mathcal{N}(\mathbf{0},\mathbf{I}),
\end{equation}
where $\bar{\alpha}_t=\prod_{i=1}^t\left(1-\beta_i\right)$, and $m_{0}$ is the ground-truth mapping. 
Afterward, following Nichol and Dhariwal~\citeyearpar{Alexander_2020_ddim}, we train a neural network $\mathcal{\epsilon}_{\theta}(\cdot)$ for the reverse denoising process, during which the initial latent variable ${m}_{T}$ is iteratively denoised following the sequence ${m}_{T-1}, {m}_{T-2}, \ldots, {m}_{0}$, as follows:
\begin{equation}
\begin{split}
\label{eq:reverse_ddim}
m_{t-1} = &\sqrt{\bar{\alpha}_{t-1}}\mathcal{\epsilon}_{\theta}(m_t, t, c_t) +\\
&\frac{\sqrt{1-\bar{\alpha}_{t-1}-\sigma_t^2}}{\sqrt{1-\bar{\alpha}_{t}}}\Bigl(m_{t} - 
\sqrt{\bar{\alpha}_{t}}\mathcal{\epsilon}_{\theta}(m_t, t,c_t)\Bigr) + \sigma_{t}z,
\end{split}
\end{equation}
where $\mathcal{\epsilon}_{\theta}(m_t, t, c_t)$ directly predicts the denoised mapping $\hat{m}_{0,t}$.

\begin{figure}[tp]
	\centering
	\includegraphics[width=0.9\linewidth]{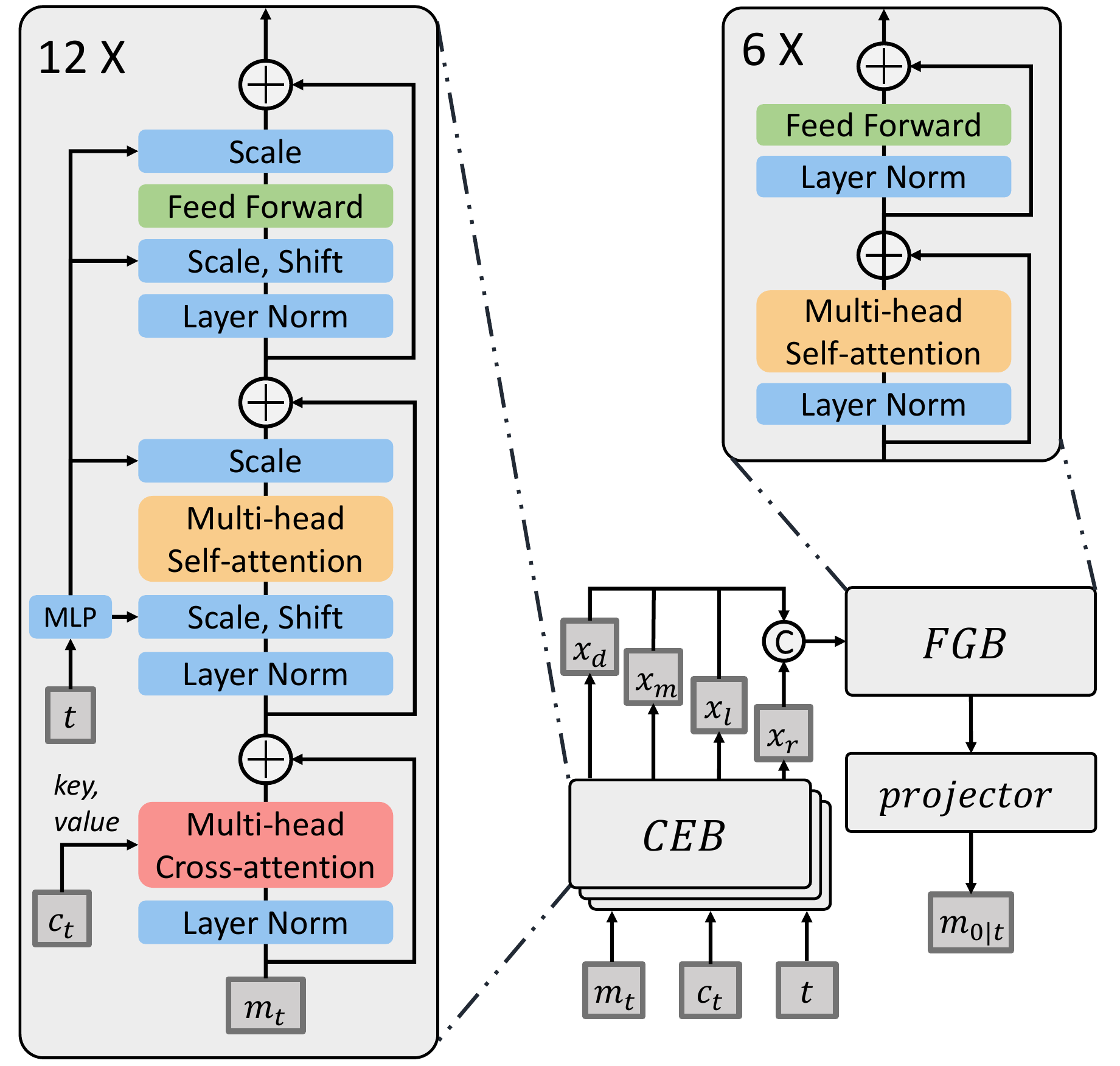}
	\caption{Detailed architectures of condition embedding blocks (CEB) and fusion generation blocks (FGB). We ignore some simple operations, such as cosine position encoding and activation functions.}
\label{fig:network}
\end{figure}

\subsubsection{Compound Conditions $c_t$.}
Following previous works~\cite{Feng2022geodocnet, li2023foreground, jiang2022RDGR}, our DvD utilizes pre-trained multiple feature extractors (MFE) to enhance document visual perception and composes these features into a compound condition for the diffusion model, denoted as $c_t=\{f_d, f_m, f_l, r_t\}$, where $f_d, f_m, f_l$ represent the features for raw document images, document foreground mask and text-lines, respectively. Note that $f_d,f_m,f_l$ are time-fixed conditions, while $r_t$ represents a time-variant condition that enables dynamic guidance in the denoising generation process. 

\subsubsection{Time-variant Condition Refinement (TVCR) Mechanism.}
As illustrated in the right region of Fig.~\ref{fig:img2}, the intermediate dewarping results reveal the visual gap from intermediate denoising states to the ideal dewarped result. To harness this information for enhanced preservation of document structures, we introduce a time-variant condition refinement mechanism within the reverse diffusion process to ensure increasingly precise document structure as the process evolves.
Specifically, we incorporate a time-variant condition $r_t=\{m_{0|t},f_{0|t}\}$ for iteratively updating $r_t$, where $m_{0|t}$ means the predicted latent variables mapping in each time-step, and $f_{0|t}$ indicates dewarped document features $f_d$ using $m_{0|t}$ by Equ.~\ref{eq:def}. Unlike the time-fixed condition applied in vanilla diffusion models, the proposed time-variant condition reflects the intermediate variable $m_{0|t}$ and local structure deformation $f_{0|t}$ in each time-step, which facilitates the model to dynamically approach a tighter evidence lower bound (ELBO)~\cite{pmlr-vae-2014} for maximizing the conditional likelihood $ \log p(m_0|\{c_t\})$.



\begin{figure*}[htp]
	\centering
	\includegraphics[width=0.98\linewidth]{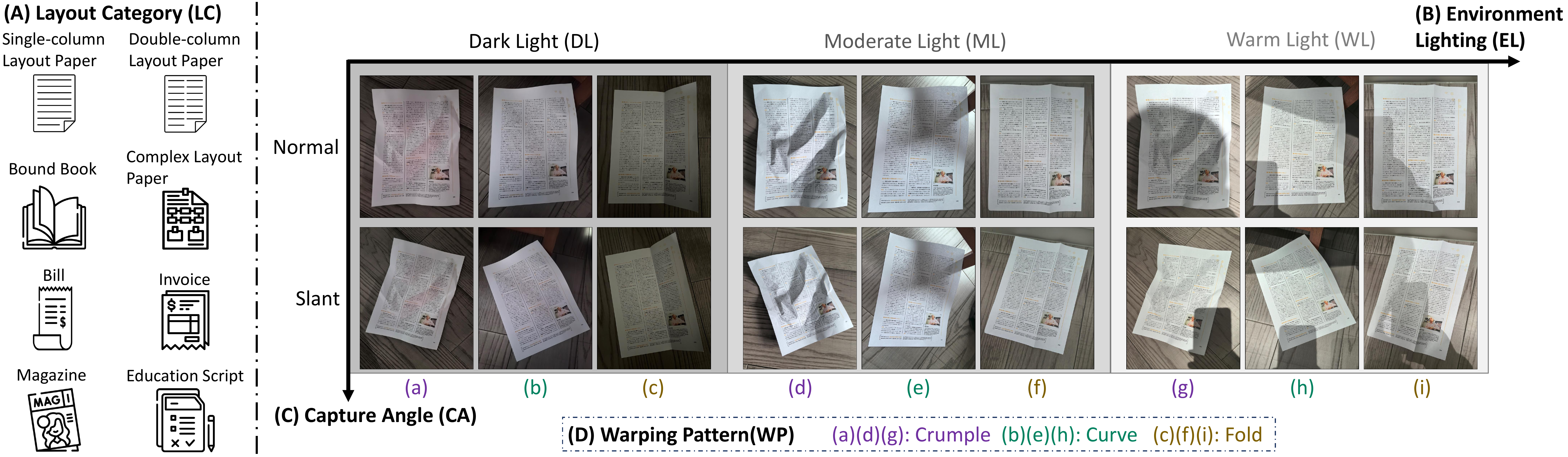}
	\caption{A sample data array visualization across 3 distinct domain spectra (A, B, C) in our AnyPhotoDoc6300 Benchmark. We aim to evaluate the dewarping capability of the model to different warping patterns (D) under well-distinguished domain combinations.}
\label{fig:benchmark_vis}
\end{figure*}

\subsection{Network Architecture}
\label{sec:method32}
In this section, we discuss the design of the network architecture in Fig.~\ref{fig:img2}, including the multiple feature extractors (\textbf{MFE}) and diffusion decoder. In MFE, we employ three parallel networks to extract features $f_d, f_m, f_l$ for raw document image, document foreground, and text lines, respectively. For document image features $f_d$, we utilize the first three blocks of VGG16~\cite{simonyan2014very} to extract features. For document foreground features $f_m$, we concatenate the last six layers of U2Net~\cite{qin_u2net_2020} as a foreground segmentation network to extract those features associated with foreground. For text lines features $f_l$, we use the last decoder layer of the UNet~\cite{ronneberger_unet_2015}, which is pre-trained for text-line segmentation. All extracted features are uniformly downsampled to $64\times64$ resolution compatible with the coordinate space.
Furthermore, as shown in Fig.~\ref{fig:network}, the diffusion decoder comprises 12 condition embedding blocks (\textbf{CEB}) and 6 fusion generation blocks (\textbf{FGB}). In each CEB, we extend the standard architecture of diffusion transformers (DiT), i.e.~DiT\_S\_2~\cite{Peebles_2023_dit} to implement both time embeddings and cross-attention conditioning.
To embed input time-steps, we adopt the same two-layer MLP in standard DiT to represent a 256-dimensional frequency embedding. To enable the condition control, we extend a multi-head cross-attention mechanism, where the compound condition $c_t$ serves as the key and value, and the noisy latent variable $m_t$ is used as the query. Specifically, we employ four parallel CEBs to decouple different conditions, including three feature conditions($f_d, f_m, f_l$) from the MFE and one time-variant condition ($r_t$). 
Subsequently, the CEB produces the hidden representations for the four conditions: $x_d$, $x_m$, $x_l$, and $x_r$.
In the FGB, the results of CEB are concatenated and fed into the FGB, which contains self-attention and feed-forward layers. Subsequently, we apply three linear layers as a projector to obtain the denoised mapping $m_0$. Next, we directly upsample $m_0$  to obtain the high-resolution mapping $M_0$, which is used to produce the final flat document image $I_0$ via Equ.~\ref{eq:def}. In Appendix B, we provide more detailed architectures.

\begin{algorithm}[tp]
	\flushleft
	\caption{DvD training for TVCR mechanism}
	\begin{algorithmic}[1] 
		\STATE  \textbf{repeat}
		\STATE  $m_0\sim q(m_0|c_t)$, $t \sim$ Uniform$(\{1,...,T\})$
		\IF{$t=T$} 
        \STATE  $r_T=\{O,O\}$, where $O$ represent all zeros for initialization 
        \ELSIF{$t<T$}
        \FOR{$t \in [T-1, t]$}
        \STATE  Sampling intermediate latent variable $m_{0|t}$ by Equ.~\ref{eq:reverse_ddim}
        \STATE  Using $m_{0|t}$ to obtain intermediate dewarped \\
                feature $f_{0|t}$ by Equ.~\ref{eq:def}
        \ENDFOR
        \STATE  $r_t=\{m_{0|t},f_{0|t}\}$
        \ENDIF
        \STATE $c_t=\{f_d,f_m,f_l,r_t\}$
		\STATE Optimize $m_{0|t}=\mathcal{\epsilon}_{\theta}(m_t, t, c_t)$ by Equ.~\ref{eq:loss}
		\STATE \textbf{until} Converged
	\end{algorithmic}
	\label{alg:training}
\end{algorithm}

\subsection{Training and Sampling}
\label{sec:method33}
\subsubsection{DvD Training Algorithm.}
Vanilla diffusion model randomly samples time-steps from a uniform distribution during the training phase~\cite{ho_2020_ddpm}, which mismatches the sequential acquisition of time-variant conditions. To solve this issue, we extend the vanilla diffusion model by integrating our TVCR mechanism, as detailed in Algorithm~\ref{alg:training}, where we implement a certain number of sampling based on current time-step, thereby obtaining the intermediate dewarping latent variables $m_{0|t}$ and the corresponding intermediate dewarped features $f_{0|t}$.

\subsubsection{Model Optimization.} During the training phase, we freeze the pre-trained weights of the foreground and text-line segmentation network from DocGeoNet~\cite{Feng2022geodocnet}, including U2Net and UNet. However, for the VGG dedicated to extracting raw document features, we jointly optimize it with the subsequent diffusion decoder. To optimize our DvD, instead of predicting noise in~\cite{ho_2020_ddpm}, we follow Luo et al.~\citeyearpar{luo2024flowdiffuser} and Nam et al.~\citeyearpar{nam_2024_diffmatch} to predict the generated object itself. Thus, the loss function is given by:
\begin{equation}
\begin{split}
\label{eq:loss}
\mathcal{L}=\mathbb{E}_{m_0\sim q(m_0|c_t), z \sim \mathcal{N}(\mathbf{0}, \mathbf{I}),t}\left[\left\|m_0-\mathcal{\epsilon}_\theta\left(m_t, t, c_t\right)\right\|^2\right],
\end{split}
\end{equation}

\subsubsection{Stochastic Sampling Property.} 
As shown in Equ.~\ref{eq:reverse_ddim}, the reverse process of DDIM~\cite{Alexander_2020_ddim} introduces $\sigma$ to inject stochasticity into the sampling trajectory. To account for this property and enhance generation stability, we implement a dual-hypothesis strategy that simultaneously generates two mappings. Afterward, we calculate the mean of the two mappings as a final result. We provide further training settings in Appendix A.


\begin{figure*}[htp]
	\centering
     \renewcommand{\arraystretch}{1}
     \resizebox{\linewidth}{!}{
	\begin{tabular}{lccccccc}
\includegraphics[width=\linewidth]{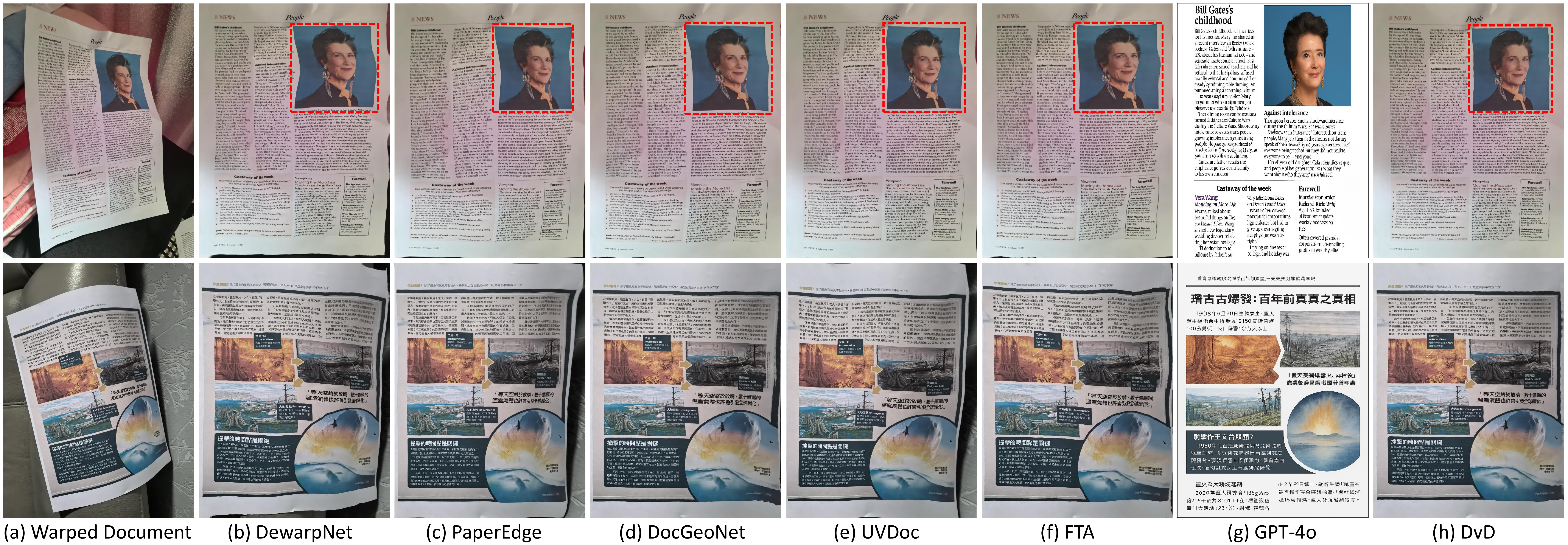}
        \\[-5pt]
        \scriptsize \hspace{3.7em}  (Input)
        \hspace{5.4em} \cite{das2019dewarpnet}
        \hspace{3.5em} \cite{RN58} 
        \hspace{3.6em} \cite{Feng2022geodocnet} 
        \hspace{2.1em} \cite{verhoeven2023neural} 
        \hspace{2.1em} \cite{li2023foreground} 
        \hspace{3.5em} (March,25th,2025)
        \hspace{5.3em} (Our)
        \\
	\end{tabular}}
	\caption{Qualitative comparisons on the AnyPhotoDoc6300 benchmark. We highlight some obvious content edges with red dotted lines. More visual comparisons can be found on the Figures-only pages after the reference.\label{fig:vis_compa1}}
\end{figure*}

\begin{table*}[htp]
\centering
\caption{Quantitative dewarping performance comparisons on the \hbox{DocUNet} benchmark dataset. \textbf{Bold} indicates the best, \underline{underline} indicates second-best. The last column shows the network size by the number of parameters (millions).}
\resizebox{\linewidth}{!}{%
	\begin{tabular}{r c c c c c c c c c c c c}
		\toprule
		Method &Venue &Training Dataset &MS-SSIM $\uparrow$ & LD $\downarrow$ & AD $\downarrow$ &  CER $\downarrow$ & ED $\downarrow$ &  MMCER $\downarrow$ & MMED $\downarrow$ & Para. \\
		\midrule
        Warped Document &- &- &0.246 & 20.51 & 1.026 & 0.595 & 1819.16 &0.576 &700.96 & - \\ \midrule
        \multicolumn{11}{c}{\cellcolor{gray!20}\textbf{Training under Non-uniform or Proprietary Dataset}} \\
		DispFlow~\cite{xie2020dewarping} &DAS'20 &DIWF &0.431 & 7.64 &0.411 & 0.446 &  1322.94 &0.887 &1339.43 & 23.6M  \\ 
		DDCP~\cite{xie2021document} &ICDAR'21 &DDCP &0.474 & 8.92 & 0.459  & 0.458 & 1335.30  &0.655 &762.28 & 13.3M \\
		PaperEdge~\cite{RN58} &SIGGRAPH'22 &Doc3D+DIW &0.472 &8.01 &0.385 & 0.407 & 1038.55 &\cellcolor{orange!45}\best{0.198} &530.27 & 36.6M  \\ 
		UVDoc~\cite{verhoeven2023neural} &SIGGRAPHA'23 &Doc3D+UVDoc &\cellcolor{orange!20}\second{0.544} &\cellcolor{orange!20}\second{6.83} &0.315 & 0.384 & 1026.91  &0.402 &615.88 & 8M \\
        LADoc~\cite{li2023ladoc} &TOG'23 &Doc3D+SP &0.523 &7.24 & 0.310 &0.395 &\cellcolor{orange!20}\second{956.27}  &0.242 &518.74 & -  \\
	DocReal~\cite{yu_docreal_2024} &WACV'24 &Doc3D+AugDoc3D &0.502 &7.00        &\cellcolor{orange!20}\second{0.284} &0.394 &1032.17 &0.336 &547.05 & -  
    \\ \midrule 	
        \multicolumn{11}{c}{\cellcolor{gray!20}\textbf{Training under Uniform Doc3D Dataset}} \\
		DewarpNet~\cite{das2019dewarpnet} &ICCV'19 &Doc3D &0.472 & 8.41 & 0.412 & 0.441 & 1158.66  &0.533 &734.84 & 86.9M\\ 
		DocTr~\cite{feng2021doctr} &MM'21 &Doc3D &0.511 & 7.77 &  0.365 & 0.403 & 1093.66 &0.432 &615.38  & 26.9M \\ 
		RDGR~\cite{jiang2022RDGR} &CVPR'22 &Doc3D&0.496 & 8.53 & 0.453 & 0.372 &994.01  &0.403 &534.97 & - \\
		Marior~\cite{jiaxin2022marior} &MM'22 &Doc3D &0.448  & 8.42 & 0.470 & 0.421 & 1131.48  &0.232 &\cellcolor{orange!45}\best{510.11} & - \\
		DocGeoNet~\cite{Feng2022geodocnet} &ECCV'22 &Doc3D &0.504  &7.69 & 0.378 & \cellcolor{orange!20}\second{0.367} & 993.08  &0.376 &627.11 & 24.8M \\ 
		FTA~\cite{li2023foreground} &ICCV'23 &Doc3D &0.494 &8.87 & 0.391 & 0.403 &1093.63  &0.355 &544.11 & 45.2M  \\
		DocScanner~\cite{feng2025docscanner} &IJCV'25 &Doc3D &0.523 &7.50 &0.333 &0.368 &1099.06  &- &- & 8.5M  \\        
		DvD &- &Doc3D&\cellcolor{orange!45}\best{0.549} &\cellcolor{orange!45}\best{6.61} & \cellcolor{orange!45}\best{0.279} &\cellcolor{orange!45}\best{0.366} &\cellcolor{orange!45}\best{928.94}  &\cellcolor{orange!20}\second{0.215} &\cellcolor{orange!20}\second{515.97}
        & 151.25M \\\bottomrule
\end{tabular}}
\label{tab:comparisontable}
\end{table*}

\section{Experiments}
\subsection{AnyPhotoDoc6300 Benchmark}
\label{sec:data}
Despite significant advances in document dewarping, the development of corresponding benchmarks lags behind. We summarized current benchmarks in Tab.~\ref{tab:unwarping_datasets}, and we can find most of datasets suffer from restricted coverage of scenarios, small size of dataset, and deficient annotations of domains, impeding a comprehensive evaluation of dewarping models. To this end, we build a new large-scale benchmark AnyPhotoDoc6300, containing 6,300 real-world photographic document pairs. In our AnyPhotoDoc6300, we provide three distinct domain annotations to enable a more fine-grained quantitative evaluation, including layout category (LC), environment lighting (EL), and capture angles (CA). 
In addition, we aim to evaluate the dewarping capability of the model on three typical warping patterns (i.e., curves, folds, and crumples) under any combination of three types of given domains.
Fig.~\ref{fig:benchmark_vis} visualizes a sample array of "Complex layout paper" (one of the eight layout categories). 
By meticulously specifying three environmental lighting and two capture angles, we can form any domain combination for the same document content. Consequently, we can obtain a fine-grained performance evaluation like Fig.~\ref{fig:vis_comparison0} to discover the underlying issues for the current dewarping model. In Appendix E, we provide more details about our AnyPhotoDoc6300 benchmark, including data collection settings and more visualizations from different layout categories. 


\subsection{Metrics}
\label{sec:metric}
\subsubsection{Feature Alignment Metrics.} Following most of previous models~\cite{ma2018docunet, das2019dewarpnet, RN58}, we adopt the MS-SSIM (Multi-scale Structural Similarity)~\cite{msssim2003}, LD (Local Distortion)~\cite{ld2017}, and AD (Aligned Distortion)~\cite{RN58} to evaluate differences between dewarped and flat ground truth. Among them, MS-SSIM focuses on perceiving similarity on luminance, contrast, and structural information. While LD and AD focus on measuring the variations of SIFT flow~\cite{liu2010sift}.

\begin{figure*}[htp]
	\centering
	\includegraphics[width=\linewidth]{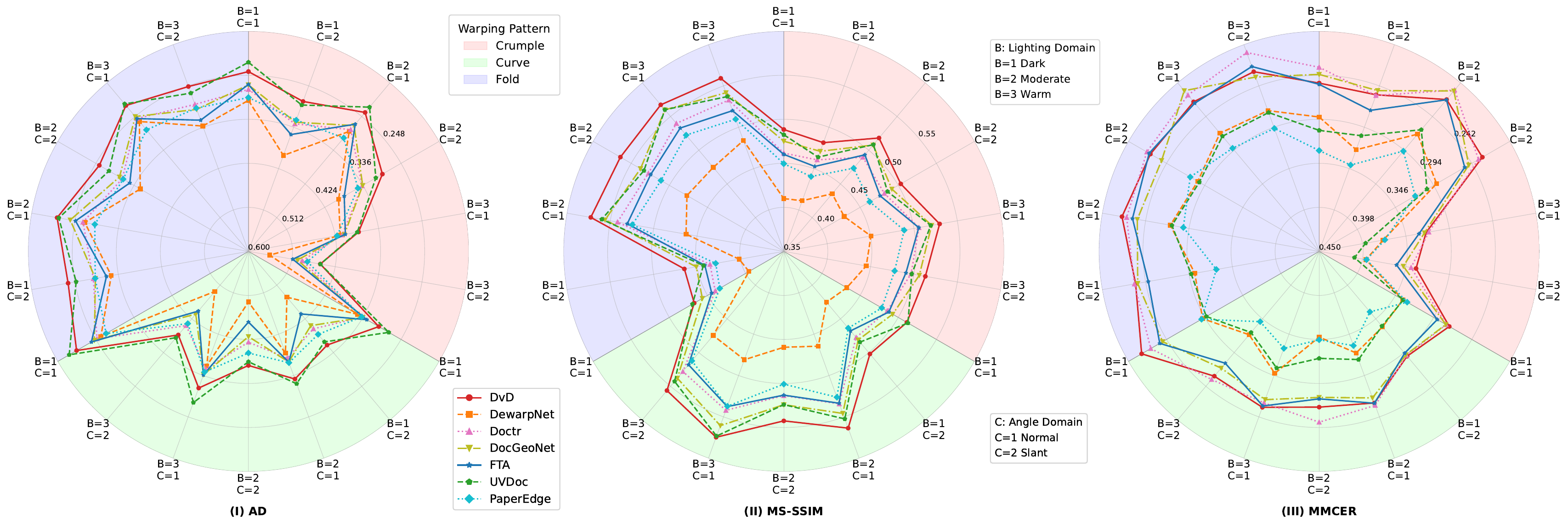}
	\caption{Quantitative dewarping performance comparisons on the \hbox{AnyPhotoDoc6600} benchmark dataset. Under the multiple Layout Pattern, we provided 18 dimensions of evaluation for AD, MS-SSIM, and MMCER. More evaluation results can be found on the Figures-only pages after the reference.}
\label{fig:vis_comparison0}
\end{figure*}

\subsubsection{OCR Metrics.} 
To evaluate the OCR readability improvement enabled by document dewarping, measuring recognition discrepancy between dewarped and flat documents has become the de facto standard in contemporary document dewarping models. Concretely, an off-the-shelf OCR engine is applied to recognize two text sequences from dewarped and flat documents, respectively. Then, ED (Edit Distance) and CER (Character Error Rate) ~\cite{levenshtein1966binary} are harnessed to quantify the degree of deviation between two sequences. 
Pioneering a metric supplement, we further extend current OCR metrics in document dewarping by replacing previous OCR engines with MLLMs.
Witnessing the superior progress of MLLMs in OCR capabilities recently~\cite{wei2024general, karmanov_eclair_2025, nassar_smoldocling_2025}, we identify that there is still no exploration about whether the dewarped document can attain equivalent readability to its flat counterpart for MLLMs.
To fill the blank, we pioneer MLLM-based OCR metrics (i.e., MMCER, MMED) to serve as a specialized supplement for prevalent ED and CER.
Specifically, given the prompt of "OCR the plain text" as a fixed instruction, we employ open-source MLLM QWen2.5-VL 7B~\cite{bai_2025_qwen25vl} to recognize all characters in both dewarped and flat documents for ED and CER calculation.



\subsection{Qualitative and Quantitative Comparison}
\subsubsection{Qualitative Comparisons.} 
\begin{figure*}[htp]
	\centering
     \renewcommand{\arraystretch}{1}
	\begin{tabular}{lccccccc}
\includegraphics[width=\linewidth]{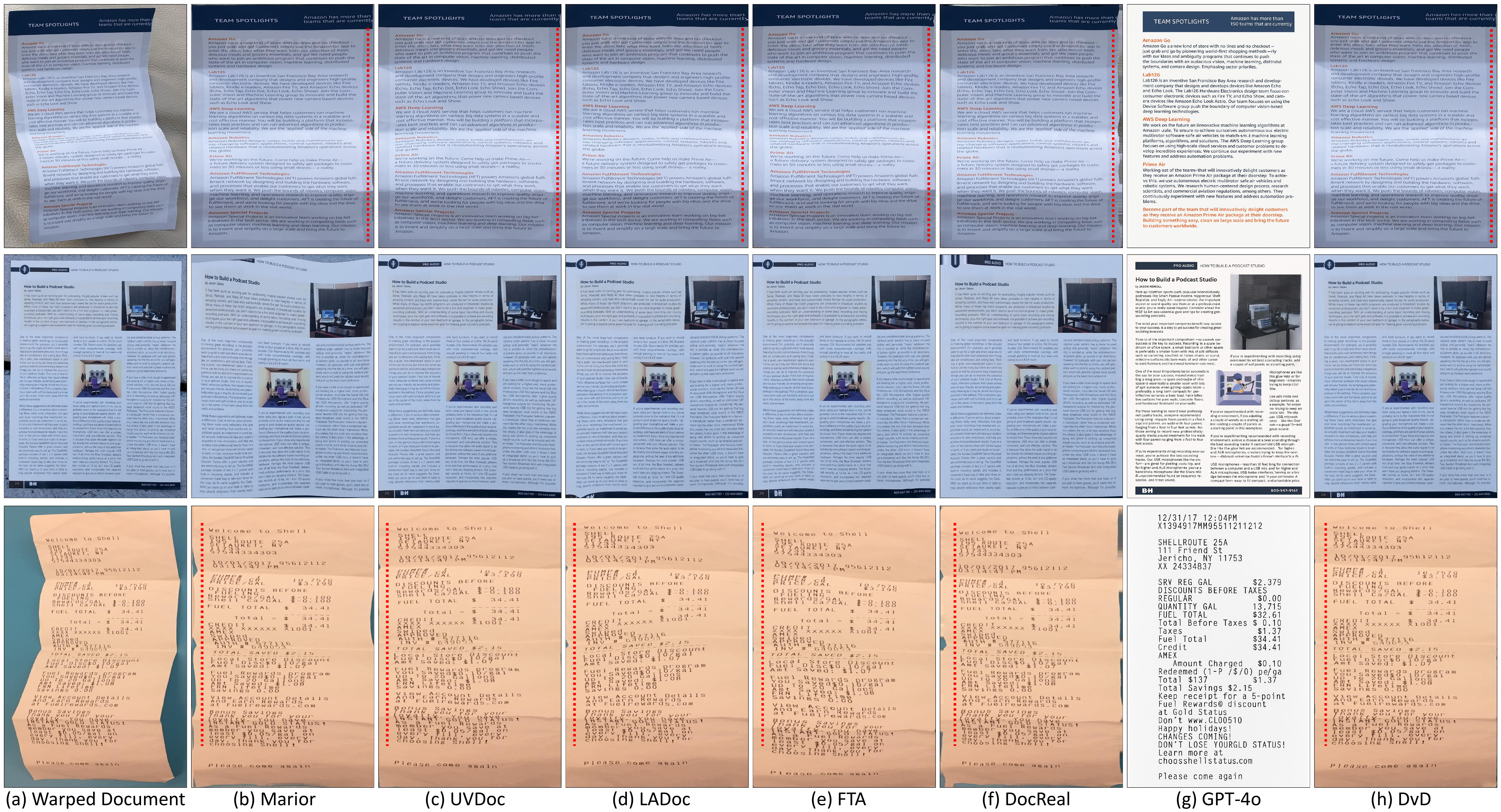}
        \\[-6.2pt]
        \scriptsize \hspace{3.9em}  (Input)
        \hspace{4.4em} \cite{jiaxin2022marior}
        \hspace{1.9em} \cite{verhoeven2023neural} 
        \hspace{2.2em} \cite{li2023ladoc} 
        \hspace{3.8em} \cite{li2023foreground} 
        \hspace{3.9em} \cite{yu_docreal_2024} 
        \hspace{3.5em} (March,25th,2025)
        \hspace{5.3em} (Our)
        \\
	\end{tabular}
	\caption{More Qualitative comparisons on the DocUNet benchmark. We highlight some obvious content edges with red dotted lines. \label{fig:vis_compa2}}
\end{figure*}

\begin{figure*}[htp]
	\centering
     \renewcommand{\arraystretch}{1}
	\begin{tabular}{lccccccc}
\includegraphics[width=\linewidth]{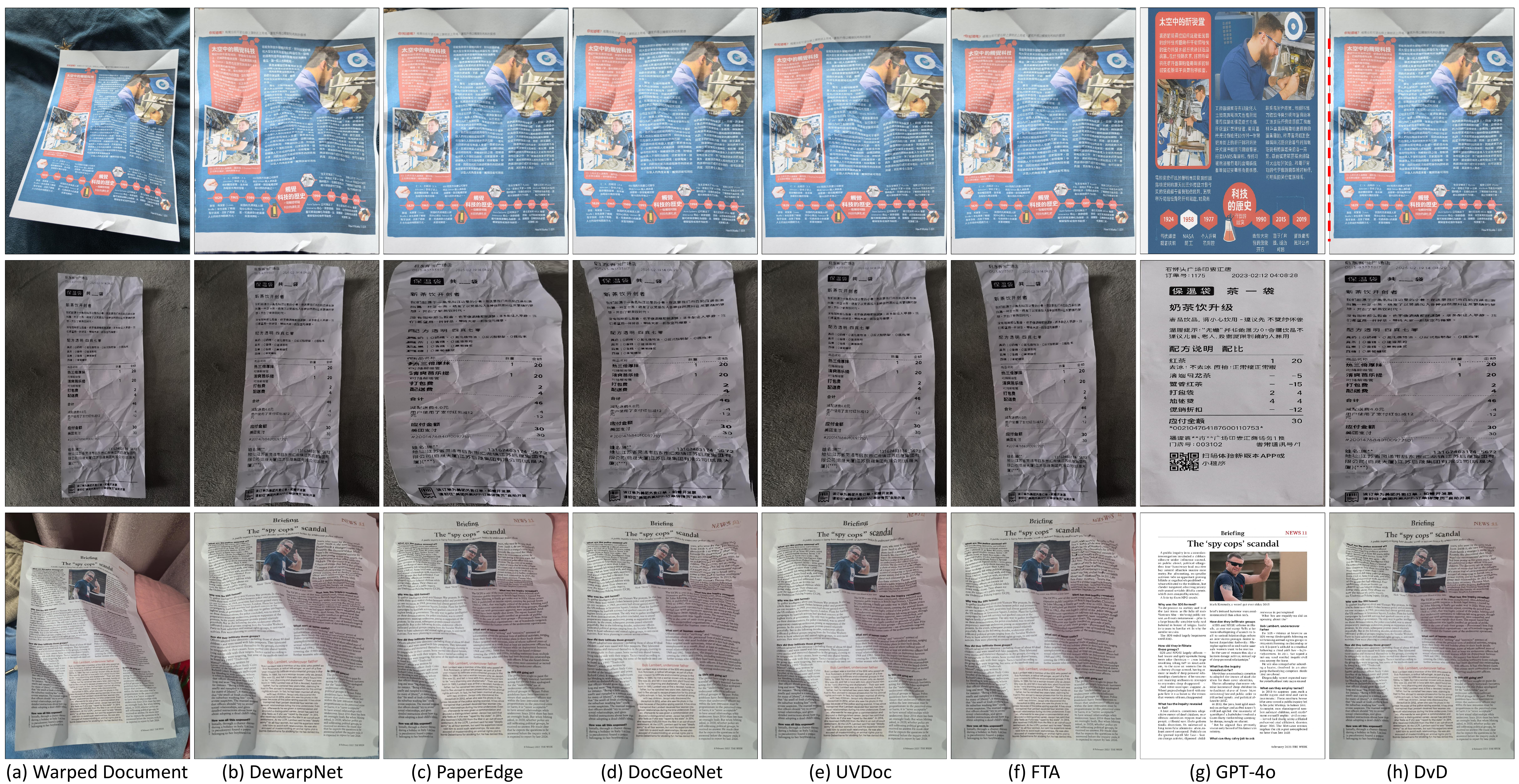}
        \\[-6.2pt]
        \scriptsize \hspace{3.6em}  (Input)
        \hspace{5.4em} \cite{das2019dewarpnet}
        \hspace{3.5em} \cite{RN58} 
        \hspace{3.7em} \cite{Feng2022geodocnet} 
        \hspace{1.9em} \cite{verhoeven2023neural} 
        \hspace{2.1em} \cite{li2023foreground} 
        \hspace{3.5em} (March,25th,2025)
        \hspace{5.3em} (Our)
        \\
	\end{tabular}
	\caption{More Qualitative comparisons on the AnyPhotoDoc6300 benchmark. We highlight some obvious content edges with red dotted lines. \label{fig:vis_compa3}}
\end{figure*}

Dewarped document results on the AnyPhotoDoc6300 and DocUNet benchmark are shown in Fig.~\ref{fig:vis_compa1}. 
In this figure, we select the five most recent open‑source dewarping models as well as GPT‑4o’s native generation model. Our prompt fed to GPT-4o is consistently given, i.e., "Please perform dewarping on this document to make it flat and clear.". On Fig.~\ref{fig:vis_compa2} and Fig.~\ref{fig:vis_compa3}, we additionally compare against publicly released inference results from non‑open‑source methods. Basically, our DvD achieves precise structure preservation in both local and overall document content. In contrast, GPT‑4o tends to produce unfaithful results that look visually clean but whose content is often chaotic. We argue this is because the image translation paradigm adopted by GPT-4o lacks explicit deformation awareness.

\subsubsection{Quantitative Comparisons.}
We compare the performance of our DvD model with previous state-of-the-art on three benchmarks, including DocUNet~\cite{ma2018docunet}, DIR300~\cite{Feng2022geodocnet}, and the proposed AnyPhotoDoc6300. 
The quantitative results on the DocUNet benchmark are shown in Tab.~\ref{tab:comparisontable}. For DIR300, we have placed the results in Appendix C. The "Warped Document" in the first row means that we simply feed the raw input image for evaluation, therefore, most of the metrics perform the worst.
Since our paper focuses on novel model designs rather than contributing high-quality data, we only train DvD on the currently most widely used Doc3D~\cite{das2019dewarpnet} dataset. For a fair comparison, we divide the current methods into two branches according to their training data. It can be seen that even though we only used the Doc3D dataset, our method still achieved the best performance on the majority of metrics.
In the upper first branch, DvD can achieve a slight overtaking, while in the lower second branch, DvD can achieve a significant superiority under a uniform Doc3D dataset. To be noted, compared with the first row, our experiments on MMCER and MMED reveal counterintuitive performance degradation in earlier dewarping models (e.g., DispFlow, DewarpNet), which we attribute to the fact that MLLMs are sensitive to resolution reduction and artifact~\cite{Li_2024_CVPR, feng_docpedia_2024}. On the other hand, compared with the first row, our DvD significantly reduces MMCER and MMED, and greatly improves the perception of photographed documents for MLLMs.

\subsubsection{Fine-grained Quantitative Comparisons.}
Fig.~\ref{fig:vis_comparison0} illustrates AD, MS‑SSIM, and MMCER as three representative metrics via the Any-PhotoDoc6300 benchmark. More metrics in different domains are also exhibited in the Fig.~\ref{fig:vis_comparison},\ref{fig:vis_comparison2},\ref{fig:vis_comparison3}, and Appendix. Our DvD comprehensively achieves superior performance across various domains, including layout, lighting, and angles. Moreover, for the first time, we can also unveil some brand-new performance comparisons for dewarping models at a fine-grained level. For instance, in Fig.~\ref{fig:vis_comparison0} (I), existing methods generally suffer from severe AD performance decline under mixed warm lighting and slant angle, especially on the warping pattern of the curves and crumple.
In Fig.~\ref{fig:vis_comparison0} (II), we observe a notable MS-SSIM drop for dark lighting documents.
In Fig.~\ref{fig:vis_comparison0} (III), warm lighting causes the most pronounced decline in OCR metrics, especially on crumpled documents.
By pinpointing these issues unconventionally, we expect to provide the research community deeper insights into dewarping model behaviors and dataset curation, for driving further performance improvements.

\subsection{Ablation Studies}
\subsubsection{Effectiveness on different Dewarping Paradigms.}
To verify the superiority of the proposed paradigm over the regression-based paradigm, we specially train another network of DvD by directly regressing the mapping. Then we can fairly compare different learning paradigms under the same network structure. As demonstrated in Tab.~\ref{tab:ablation1}, The DvD baseline model trained using the regression-based paradigm has led to a general decline in performance, which emphasizes the effectiveness of our mapping generation paradigm for obtaining a more precise structure preservation.

\subsubsection{Component Analysis of Compound Conditions.}
Tab.~\ref{tab:ablation2} demonstrates three types of compound condition $c_t$ configurations. We can see that only using the raw document feature $f_d$ does not obtain satisfactory performance, which is then improved by adding the document foreground $f_m$ and text-lines $f_l$. Finally, adding a time-variant condition $r_t$ further boosts the performance, obviously on all metrics. All of these verify the complementarity of those compound conditions.  

\begin{table}[tp]
\centering\caption{Ablation study on different learning paradigms.}
\vspace{-5pt}
\resizebox{\linewidth}{!}{
   \begin{tabular}{lccccc}
        \toprule
         Learning paradigms &MS-SSIM $\uparrow$ & LD $\downarrow$ & AD $\downarrow$ &  CER $\downarrow$ & ED $\downarrow$
         \\
        \midrule
DvD (Mapping Regression)  & 0.487 &7.85 &0.482 &0.410 &1084.67 \\
DvD (Mapping Generation) &\textbf{0.549} &\textbf{6.61} &\textbf{0.279} &\textbf{0.366} &\textbf{928.94} \\
        \bottomrule
\end{tabular}}
\label{tab:ablation1}
\end{table}

\begin{table}[tp]
\centering\caption{Ablation study for different conditions components}
\vspace{-5pt}
\resizebox{0.88\linewidth}{!}{
\begin{tabular}{ccc|ccccc}
\toprule
\multicolumn{3}{c|}{Conditions $c_t$} & \multicolumn{5}{c}{ Experimental Results } \\ 
 $f_d$ & $f_m,f_l$  & $r_t$    &MS-SSIM $\uparrow$ & LD$\downarrow$ & AD $\downarrow$ & CER $\downarrow$ & ED $\downarrow$\\
\midrule
$\checkmark$ & &               &0.409  & 8.18 &0.332 & 0.427 & 1184.52\\
$\checkmark$ &$\checkmark$ &   &0.519 & 7.12 & 0.323 & 0.408 & 1046.71\\
$\checkmark$ &$\checkmark$ &$\checkmark$ &\textbf{0.549} &\textbf{6.61} &\textbf{0.279} &\textbf{0.366} &\textbf{928.94}\\
\bottomrule
\hline
\end{tabular}}
\label{tab:ablation2}
\end{table}

\begin{table}[tp]
\caption{Ablations on sampling step, performance, and time consumption.}
\vspace{-5pt}
\centering
\resizebox{0.92\linewidth}{!}{
\color{black}{
   \begin{tabular}{l|cccccc}
        \toprule
        Steps &MS-SSIM $\uparrow$ & LD$\downarrow$ & AD $\downarrow$ & CER $\downarrow$ & ED $\downarrow$ & Time $\downarrow$ \\ \midrule
        1  &0.420 &9.86 &0.501 &0.875 &1952.54 &\textbf{0.21}\\ 
        3   &\textbf{0.549} &\textbf{6.61} &\textbf{0.279} &\textbf{0.366} &\textbf{928.94} & 0.59\\ 
        5   &0.537 &6.60 &0.281 &0.372 &956.46 &1.06\\
        50  &0.475 &7.56 &0.467 &0.441 &1261.43 &10.32\\
        \bottomrule
\end{tabular}}}
\label{tab:ablation3} 
\end{table}

\begin{table}[tp]
\caption{Ablations study for different latent size.}
\vspace{-5pt}
\centering
\resizebox{0.82\linewidth}{!}{
\color{black}{
   \begin{tabular}{c|ccccc}
        \toprule
        Size &MS-SSIM $\uparrow$ & LD$\downarrow$ & AD $\downarrow$ & CER $\downarrow$ & ED $\downarrow$  \\ \midrule
        $16\times16$  &0.432 &10.63 &0.462 &0.451 &1343.87 \\ 
        $32\times32$   &0.492 &7.69 &0.370 &0.385 &1023.28 \\ 
        $64\times64$  &0.549 &\textbf{6.61} &0.279 &\textbf{0.366} &\textbf{928.94} \\
        $128\times128$  &\textbf{0.551}&6.62&\textbf{0.276}&0.376&940.43\\
        \bottomrule
\end{tabular}}}
\label{tab:ablation4} 
\end{table}

\subsubsection{Computational Efficiency Analysis.}
Tab.~\ref{tab:ablation3} illustrates a trade-off between performance and time consumption, driven by different diffusion denoising steps. The time unit here refers to the average elapsed seconds per image we take to infer the DocUNet~\cite{ma2018docunet} benchmark. As the number of sampling steps increases from 1 to 3, the model’s performance improves substantially. However, beyond 3 steps, performance largely plateaus or even slightly declines. We consider that this might be due to excessive steps that could accumulate errors over time, causing generated samples to deviate from the real distribution. At the same time, increasing the step count incurs a linearly growing time cost. In our experiments, we set 3 as the optimal step setting for a trade-off between performance and time consumption.

\subsubsection{Size Analysis for Latent Space.}
Tab.~\ref{tab:ablation4} presents the performance of the DvD model under different latent space resolutions. When the resolution is too small ($16\times16$), the model performs poorly. We attribute this to the fact that such a limited latent space sacrifices overmuch warping semantics, making it difficult to accurately represent the high-resolution (i.e., $2000\times3000$) backward mapping $M_0$. Empirically, we find that a moderate resolution of $64\times64$ is sufficient to provide warping semantics. Further increasing the resolution to $128\times128$ yields no significant performance gain.


\subsubsection{Limitations.} 
\begin{figure*}[htp]
	\centering
	\includegraphics[width=\linewidth]{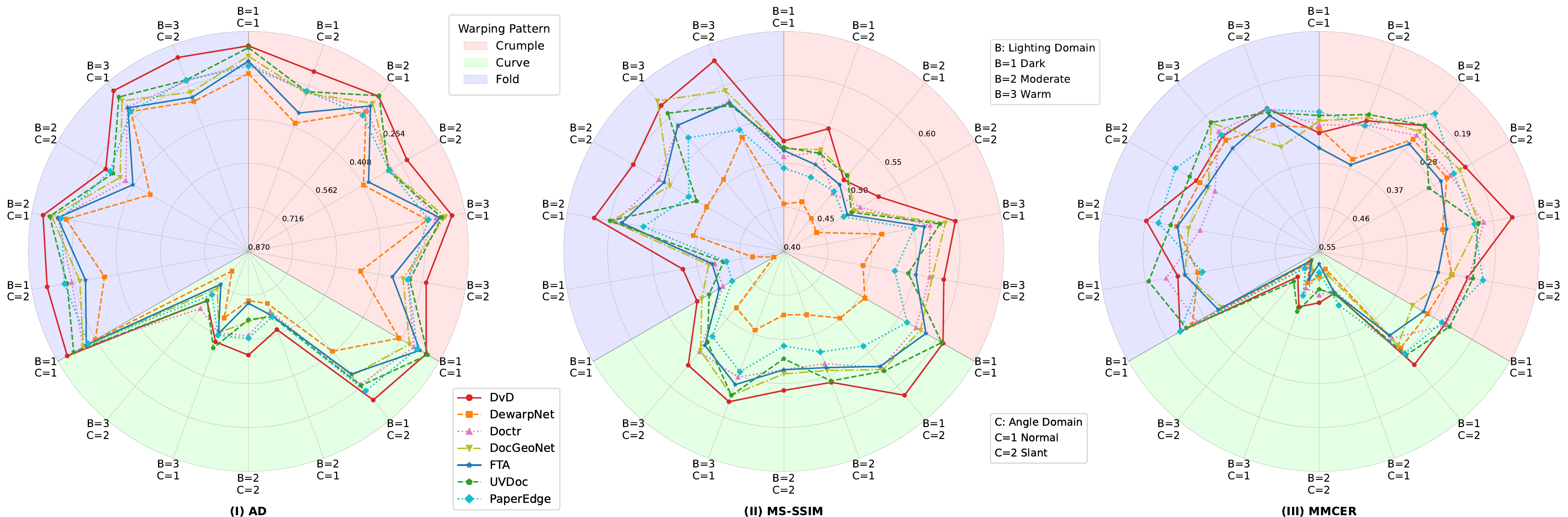}
	\caption{Quantitative dewarping performance comparisons on the \hbox{AnyPhotoDoc6600} benchmark dataset. Under the fixed "Two-column Paper" Layout Pattern, we provided 18 dimensions of evaluation for AD, MS-SSIM, and MMCER. More evaluation results can be found on the Figures-only pages after the reference.}
\label{fig:vis_comparison}
\end{figure*}

\begin{figure*}[tp]
	\centering
	\includegraphics[width=0.95\linewidth]{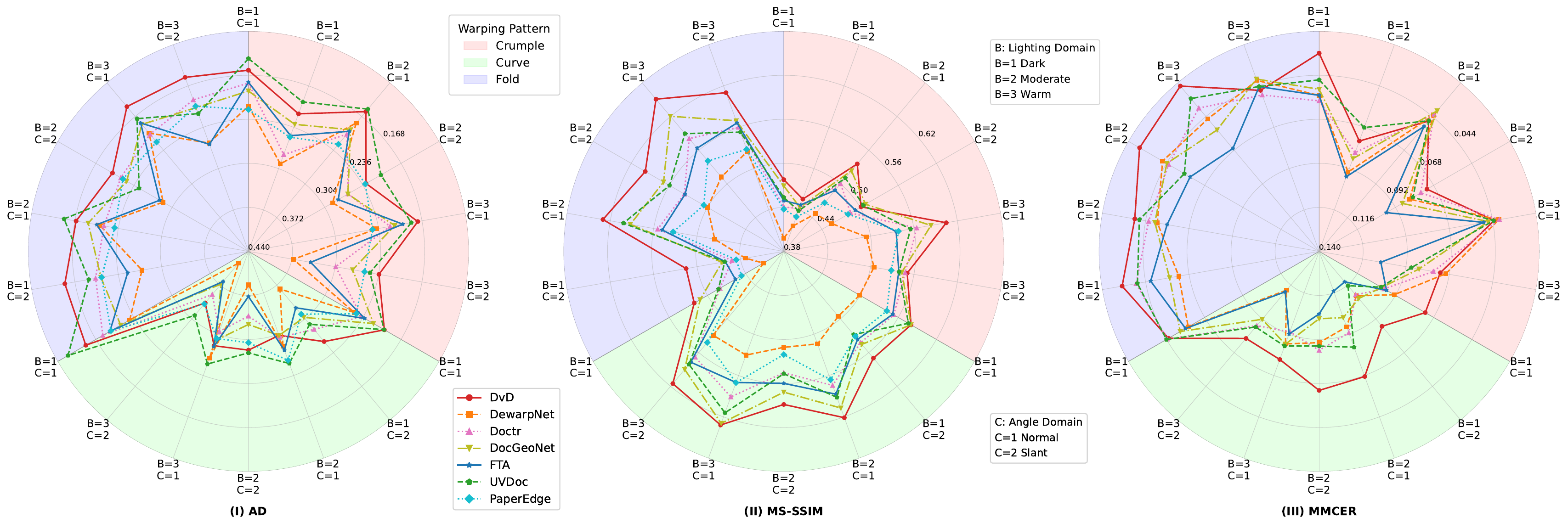}
	\caption{More Quantitative dewarping performance comparisons on the \hbox{AnyPhotoDoc6600} benchmark dataset. Under the fixed "Single-column Paper" Layout Pattern, we provided 18 dimensions of evaluation for AD, MS-SSIM, and MMCER. }
\label{fig:vis_comparison2}
\end{figure*}

\begin{figure*}[tp]
	\centering
	\includegraphics[width=0.95\linewidth]{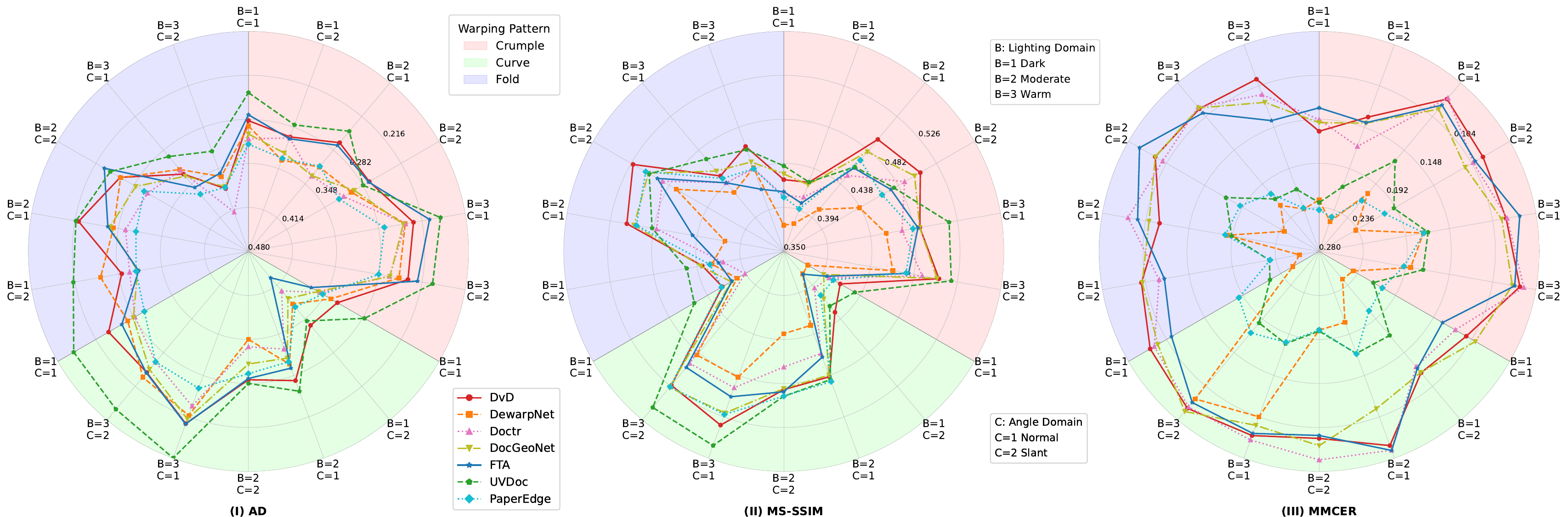}
	\caption{More Quantitative dewarping performance comparisons on the \hbox{AnyPhotoDoc6600} benchmark dataset. Under the fixed "Magazine" Layout Pattern, we provided 18 dimensions of evaluation for AD, MS-SSIM, and MMCER.}
\label{fig:vis_comparison3}
\end{figure*}

Our method still has two limitations. (1) Slow training: Unlike directly selecting the denoising time-step in vanilla DDIM, training with the TVCR mechanism requires sampling a few steps per iteration, causing a slower training speed.
(2) Limited Generalization on unseen document types: Diffusion models excel at generating samples that conform to the training data distribution. Since the training set Doc3D~\cite{das2019dewarpnet} contains many academic papers and magazines, DvD shows superiority on these seen document types (cf. Fig.~\ref{fig:vis_comparison},\ref{fig:vis_comparison2},\ref{fig:vis_comparison3}). However, DvD's improvements on unseen types (Invoices/Education scripts) are negligible (cf. Fig.~15,16 in Appendix).


\section{Conclusion}
This paper unleashes a novel mapping generation paradigm for the document dewarping task by reformulating it as a coordinate-based denoising diffusion framework.
To the best of our knowledge, this is the first attempt to explore the viability of dewarping a document using the diffusion model, where we tailor a coordinate‑level denoising strategy and a time-variant condition refinement (TVCR) mechanism, enabling precise preservation of document structures.
To foster a fine-grained evaluation of dewarping models, we also build a new photographic document dewarping benchmark, AnyPhotoDoc 6300, which is large-scale in size, covers multiple scenarios, and provides detailed domain annotations.
Findings and insights from our experiments are poised to substantially advance photographic document processing and further impact a broad spectrum of graphics applications.

\begin{acks}
The work was partially supported by the following: National Natural Science Foundation of China under No. 92370119, 62436009, 62276258 and 62376113, XJTLU Funding REF-22-01-002.
\end{acks}


\newpage
\bibliographystyle{ACM-Reference-Format}
\bibliography{reference}

\end{document}